\pdfoutput=1
\documentclass[11pt]{article}

\usepackage[final]{acl}

\usepackage{multirow}
\usepackage{tabularx}
\usepackage{graphicx}
\usepackage{subfigure}

\usepackage{times}
\usepackage{latexsym}

\usepackage[T1]{fontenc}
\usepackage[utf8]{inputenc}

\usepackage{microtype}

\usepackage{inconsolata}

\usepackage{amsmath,amssymb,amsfonts}
\usepackage{graphicx}
\usepackage{textcomp}
\usepackage{xcolor}
\usepackage{booktabs}
\usepackage{multirow}
\usepackage{xspace}
\usepackage{balance}
\usepackage{bm}
\usepackage{url}
\usepackage{hyperref}
\usepackage{marvosym}
\usepackage{multicol}
\usepackage{xspace}
\usepackage[flushleft]{threeparttable}
\usepackage{adjustbox}
\usepackage{graphicx}
\usepackage{subcaption}
\usepackage{enumitem}
\usepackage[ruled,linesnumbered]{algorithm2e}

\newcommand{\ignore}[1]{}

\newcommand{\ours}{DEA-SQL }
\newcommand{\nours}{DEA-SQL}

\newcommand{\luxc}[1]{\textcolor{black}{#1}}
\newcommand{\xieyzh}[1]{\textcolor{black}{#1}}

\author{Yuanzhen Xie$^{2}$ \, Xinzhou Jin$^{1}$\, Tao Xie$^{2}$ \, Mingxiong Lin$^{2}$\, Liang Chen$^{1}$\, \\ \textbf{Chenyun Yu$^{1}$ \thanks{Chenyun Yu is the corresponding author}\, Lei Cheng$^{2}$\, Chengxiang Zhuo$^{2}$\,  Bo Hu$^{2}$\, Zang Li$^{2}$ }\\
$^1$Sun Yat-sen University\\
$^2$Platform and Content Group, Tencent\\
\texttt{\{xieyzh3, taoxie168\}@gmail.com}\\
\texttt{jinxzh5@mail2.sysu.edu.cn}\\
\texttt{\{chenliang6, yuchy35\}@mail.sysu.edu.cn}\\
\texttt{\{matrixmxlin, raycheng, felixzhuo, harryyfhu, gavinzli\}@tencent.com}}

\title{Decomposition for Enhancing Attention: Improving LLM-based Text-to-SQL through Workflow Paradigm}
\begin{document}
\maketitle
\begin{abstract}

In-context learning of large-language models (LLMs) has achieved remarkable success in the field of natural language processing, while extensive case studies reveal that the single-step chain-of-thought prompting approach faces challenges such as attention diffusion and inadequate performance in complex tasks like text-to-SQL. To improve the contextual learning capabilities of LLMs in text-to-SQL, a workflow paradigm method is proposed, aiming to enhance the attention and problem-solving scope of LLMs through decomposition. Specifically, the information determination module for eliminating redundant information and the brand-new prompt structure based on problem classification greatly enhance the model's attention. Additionally, the inclusion of self-correction and active learning modules greatly expands the problem-solving scope of LLMs, hence improving the upper limit of LLM-based approaches. Extensive experiments conducted on three datasets demonstrate that our approach outperforms other methods by a significant margin. About 2-3 percentage point improvements compared to the existing baseline on the Spider Dev, Spider-Realistic, and Bird Dev datasets and new SOTA results on the Spider Test dataset are achieved. Our code is available on GitHub: \url{https://github.com/FlyingFeather/DEA-SQL}.

\end{abstract}

\section{Introduction}
Many datasets are stored in tabular form, utilizing relational databases. Extracting data from such sources typically requires skilled professionals to access it through manually crafted structured query language (SQL) queries. This limitation restricts the accessibility of ubiquitous relational data to a broader range of non-technical users. The text-to-SQL parsing~\citep{finegan2018improving, yu2018spider, qin2022survey} task aims to bridge this gap by translating natural language questions (NLQ) into SQL query statements given a database schema without additional knowledge. This significantly reduces the barrier to database access, thereby garnering increasing attention from both academia and industry.

Early works on text-to-SQL parsing in the database community~\citep{wang2020rat, scholak2021picard, hui2022s2sql} have made significant strides. A dominant approach involves collecting labeled data and training models through supervised learning. While effective, this method requires a large amount of labeled training data, which entails high costs in both annotating SQL queries and conducting model training.
As an alternative to supervised learning, in-context learning~\citep{dong2022survey} represents an emerging capability of large language models (LLMs), reducing the reliance on extensive datasets. With just a few examples, in-context learning enables LLMs to demonstrate performance comparable to or even surpassing fully supervised models across many natural language processing (NLP) tasks, such as sentiment analysis, machine translation and natural language inference~\citep{liang2023few, xu2023paradigm, wei2022chain}.

In-context learning has also shown promising results in text-to-SQL parsing. Existing in-context learning methods~\citep{tai2023exploring, chang2023prompt} often employ techniques like chain-of-thought (COT), using prompts to accomplish complex text-to-SQL tasks. We conduct single-step COT experiments using GPT-4 on the Spider dataset and find that while COT does increase LLMs' problem-solving scope (the correct solution produced by the LLMs) to some extent, however, the performance of these single-step prompts is limited.
For example, instructions added within a lengthy prompt text might be disregarded due to the LLM's attention being diluted by the excessive length of the text. Additionally, improving the prompt text often results in a seesaw effect, making it difficult to improve the overall performance as it is not universal for different types of problems.

Unlike traditional NLP tasks, text-to-SQL, which involves converting natural language to structured language, is more challenging. The DIN-SQL~\cite{pourreza2023dinsql} method provides a new direction and achieves some success by decomposing steps and incorporating universal self-correction. 
However, this approach lacks targeted steps for specific error types, which restricts the effective expansion of LLMs' problem-solving scope. Also with its fixed few-shot learning mechanism, it hampers the adaptability and generalizability of LLMs.

In this paper, we propose a workflow paradigm method for decomposing and enhancing attention based on few-shot. This method draws on human thinking patterns, adheres to the principle of making subtasks as simple as possible, and reduces irrelevant information in each step to specifically enhance the solvable scope of LLM and improve the attention of LLM to enhance their performance. It consists of five sub-modules: an \textbf{Information Determination} module that focuses on attention by reducing interference information through a two-stage method; a \textbf{Classification \& Hint} module that solves different problems that cannot be generalized by simply providing different simple prompts; a few-shot \textbf{SQL Generation} module based on question template retrieval; a \textbf{Self-Correction} module based on error summarization; and an \textbf{Active Learning} module that expands the model's capabilities based on error cases. To reduce consumption, we minimize the number of few-shots required for each step, with the second stage of field filtering and the self-correction step being zero-shot and the SQL generation module outperforming existing few-shot-based baselines in a zero-shot environment. 
\luxc{It has been found that the workflow prompting paradigm is more effective in improving overall performance compared to single-step prompting. In the experiments, execution accuracies of 85.4 and 81.5 were achieved on the Spider-dev and Spider-Realistic datasets, respectively, surpassing existing models and in-context learning schemes.}

Our contributions can be summarized as:
1) Propose a workflow paradigm solution to boost the attention of LLMs for complex problems as an example for text-to-SQL tasks;
2) Design a two-stage information filtering module to curtail irrelevant information to enhance the attention of LLMs, while adapting realistic questions with different questioning styles, which performs better on datasets that are closer to realistic questioning styles;
3) Propose a new prompt structure for text-to-SQL tasks. Categorize the problems and use different prompt patterns for different types of problems, presenting the key information to the model in a more explicit way to better improve the performance of the model;
4) The integration of LLMs for self-correction and active learning further improves the model.

\section{Methodology}
\begin{figure*}[ht]
  \centering
  \includegraphics[width=0.9\linewidth]{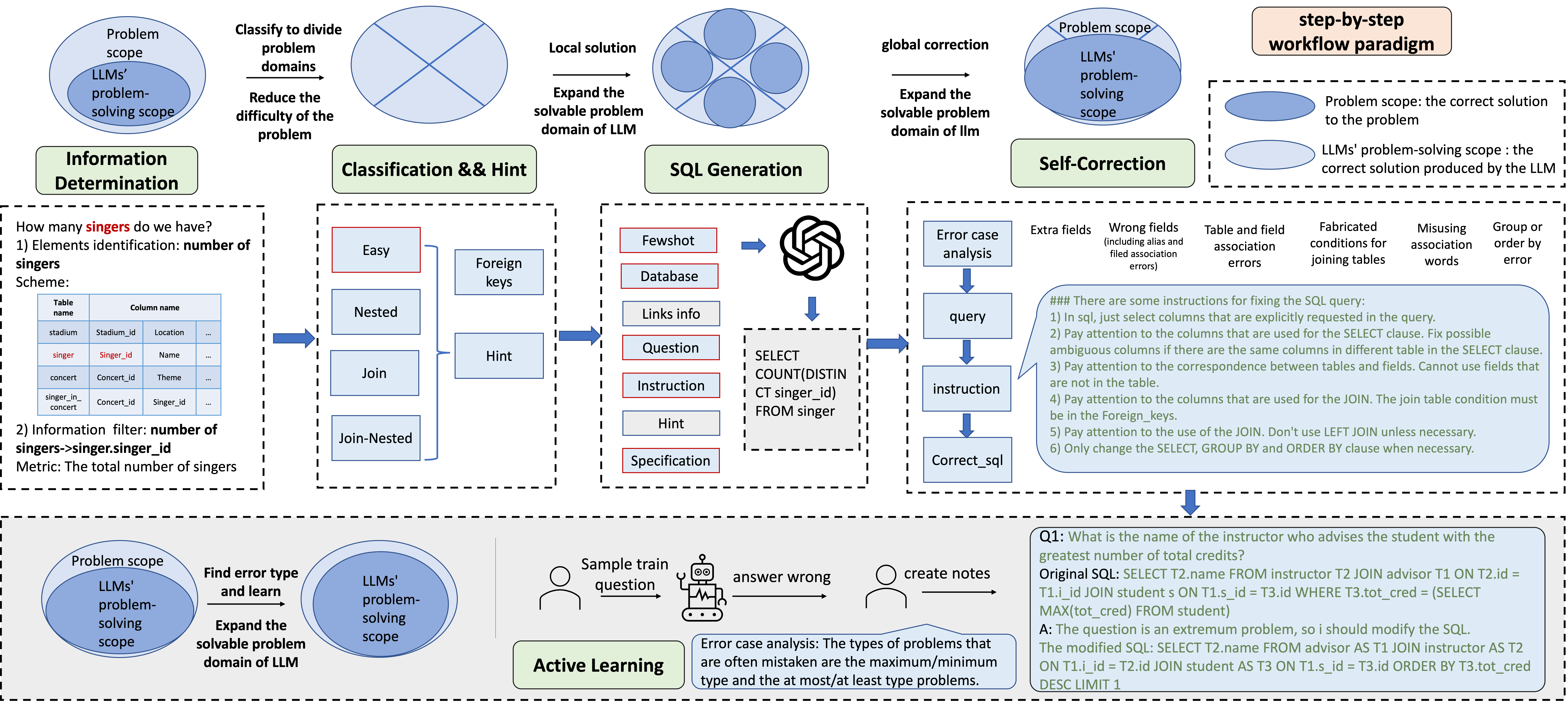}
  \caption{ The overall structure of the \ours model.}
  \label{fig:sys}
\end{figure*}
The emergence of chain-of-thought based on prompt engineering has stimulated the ability threshold of large language models(LLMs) to shine. Although prompt engineering has achieved some success in major domains, relying solely on the stimulation of prompt engineering is not enough for LLMs to learn to solve complex tasks. On the contrary, this approach of stacking a large amount of text content poses several major drawbacks: 1) If the LLM focuses on a lot of points at once, its attention becomes diluted and it is difficult to cover all the points, so the effect may be reduced based on not concocting the prompts attentively. 2) It is more difficult to get LLMs to focus on the specific issues you raise within a large amount of text.

When faced with complex problems, people's solutions are usually to continuously try to break them down into multiple steps, ultimately coming up with a relatively complete process methodology to improve quality and efficiency.
Inspired by this idea, we attempt to approach it from the perspective of workflow, breaking down the work into finer details to \textcolor{black}{address the challenges mentioned in the introduction}. Each time, we let the LLMs focus on a single atomic task, adhering to the principle of keeping the task as simple as possible and the information as streamlined as possible. This allows the LLMs' attention to be concentrated, thereby enhancing their comprehension ability.

In the text-to-SQL task, the common solution process can be summarized as follows: 1) Determine the necessary database information; 2) Identify the query type of the problem; 3) Consider the problem-solving approach based on the question type and write the corresponding SQL; 4) Perform a preliminary self-check on the SQL; 5) Recall past mistake examples, check if the current answer has the same error points, and avoid repeating the same mistakes.
Thus, based on the idea that \textbf{d}ecomposition for \textbf{e}nhancing \textbf{a}ttention, we propose the workflow paradigm method named \ours with five modules as shown in Figure \ref{fig:sys}: Information determination, Classification \& Hint, SQL generation, Self-correction, Active learning. 
Some modules like information determination can be well-solved using traditional techniques.
We choose to explore the comprehensive capabilities of LLMs at each step to demonstrate the effectiveness of the attention-focused workflow paradigm in in-context learning. Thus, each step is implemented using the LLM based on prompt engineering.

\subsection{Information Determination}
The information determination step is primarily responsible for identifying the schema of the specific database needed for the problem. 
Some studies~\cite{lei2020re, pourreza2023dinsql} have demonstrated that schema linking facilitates cross-domain generalization and synthesis of complex queries.
A simple starting point for this step is that giving too much unwieldy information to the LLMs can overly interfere with their ability to comprehend it.
By minimizing such interference, we can potentially improve the attention and effectiveness of the LLMs.
Specifically, determine the elemental information about the tables and columns that the problem needs to use before determining the computational logic for the specific SQL, and pare down any other extraneous table information.

Common techniques for determining table and column information typically rely on one-step prompts and few-shots approaches \cite{dong2023c3, pourreza2023dinsql}. However, due to the more open-ended nature of the questions being asked, these methods often struggle to capture all the key vocabulary, rendering them relatively ineffective.
To address the issue of instability in LLMs caused by changes in the problem, we propose a two-stage approach that differs from existing methods: we first identify the problem elements(elements identification) and then go through them to select the required vocabulary to further refine the relatively complex problem (information filter) as shown in Figure \ref{fig:sys}.
The primary issues requiring attention can also be identified during this stage, allowing the LLMs to contemplate these concerns during the SQL generation phase.

\subsection{Classification \& Hint}
\begin{figure*}[ht]
  \centering
  \includegraphics[width=0.9\linewidth]{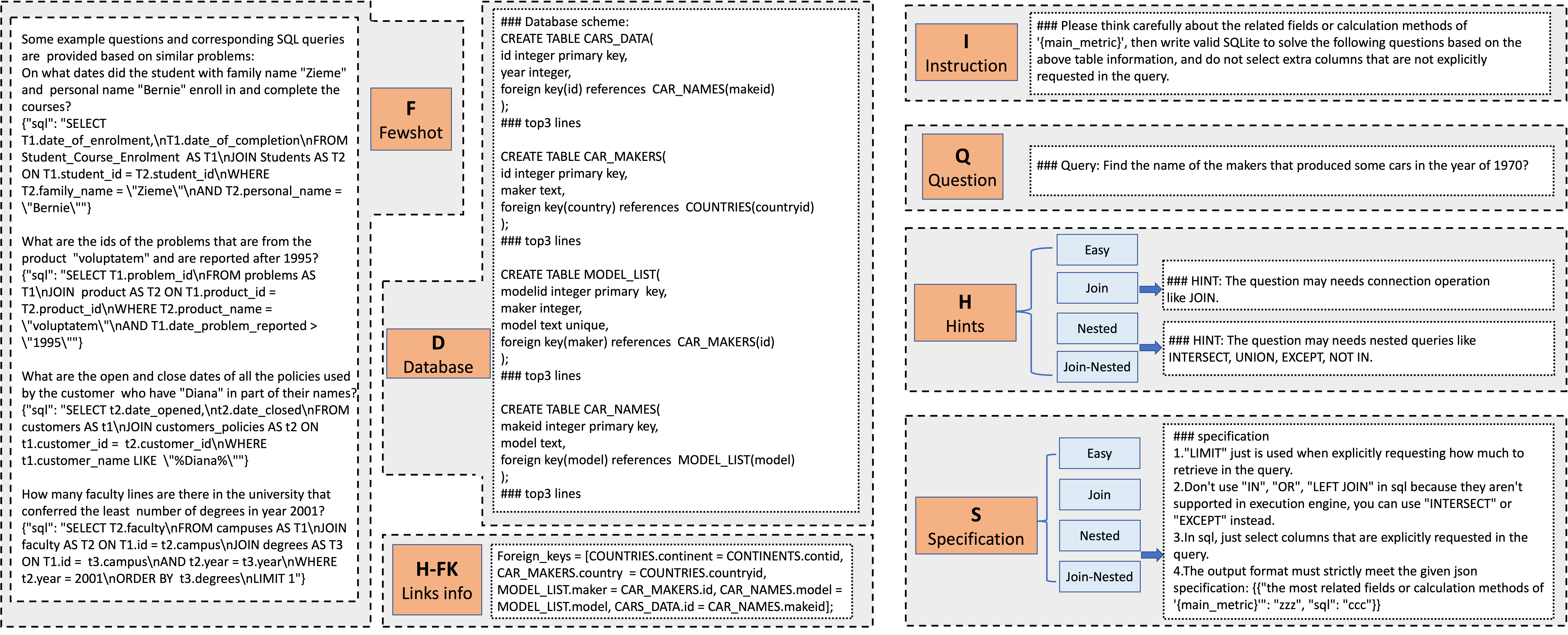}
  \caption{ \ours model's prompt structure in SQL Generation module.}
  \label{fig:str}
\end{figure*}

In the error analysis of directly using the LLM and inspired by DIN-SQL~\cite{pourreza2023dinsql}, it is found that it has weaker support for nested queries and joint problems. The accuracy of distinguishing between nesting and joining when directly writing SQL is relatively low. It implements the join problem using "IN" or other nested queries.
In addition, for complex multi-level joint problems, it is challenging to comprehend the correct join conditions at a glance when confronted with a plethora of table creation information. This makes LLMs easy to envision or employ incorrect join conditions.

To address these issues, the classification \& hint module is proposed. We divide the problems into four major categories based on whether the SQL requires nested subqueries and whether it requires joining tables: easy, join, nested, and join-nested problems.
The "easy" category refers to problems that do not require joining tables or utilizing nested queries. The "join" category pertains to problems that necessitate joining tables but not nested queries. The "nested" category encompasses problems that do not require joining tables but do demand nested queries. Finally, the "join-nested" category includes problems that call for both joining tables and employing nested queries.
Subsequently, in the following SQL generation step, we add special prompt words for different problem types to remind it to pay attention to the relevant problem types as shown in Figure \ref{fig:sys}.

Since SQL is a language for structured databases, there is a certain gap between it and natural language. As query problems grow more complex, there is a need for additional information to enhance the precision of the LLMs. To help the LLM focus its attention, we provide additional information about join conditions in the form of a list structure. 
Comparing the Database and Link info in Figure \ref{fig:sys}, it is obvious that the Link information is more intuitive for humans or models to select the correct join condition.
In addition, different hints (H) are provided in SQL generation based on the type of question as shown in Figure \ref{fig:str} for the Hints.

\subsection{SQL Generation}
Different question types necessitate various focal points. In accordance with the classification of questions, we incorporate distinct prompts for each category to emphasize the aspects that demand particular attention within that question type.
The overall question prompt follows the format $<F, D, H-FK, I, Q, H, S>$ shown as Figure \ref{fig:str}, and simple questions do not require a concatenation condition (i.e., simple questions follow the format $<F, D, Q, H, S>$), where $F$ represents few-shots that are optional, $D$ denotes database, $H-FK$ represents the foreign keys that are optional, $I$ indicates command information that describes the problem requirements, $Q$ represents the question, $H$ denotes the hints depending on the question type and $S$ denotes the specification depending on the question type.

Complex problems, which may be relatively less seen from the model's perspective, cannot be effectively solved by merely modifying the prompt. To stimulate the model's ability to solve such problems, we use specific examples as prompts.
Experiments have found that LLMs are highly sensitive to sample selection, and choosing irrelevant samples may even produce negative effects. This concept is straightforward to grasp: absorbing too many solutions to unrelated problems can distract the model, making it more challenging to determine the correct method for problem-solving.

In the few-shot version, we also categorize the few-shot library according to problem type. Intuitively, examples identified based on problem type are more likely to fulfill the requirements of finding the most relevant problem. We establish various retrieval methods, which can be divided into two main types: full retrieval and retrieval by problem type. In the latter, there are three subtypes: random, question similarity (ques\_sim), and template similarity (tem\_sim).
Relying solely on question similarity can often result in retrieving questions that aren't particularly similar. Template similarity, on the other hand, begins by obtaining the retrieval question paradigm and mitigates the effect of question diversity. To obtain the retrieval question paradigm, key entities for word selection are concealed to generate the question skeleton (template). This approach can effectively minimize the impact of different entities in similar questions, or identical entities in dissimilar questions.

\subsection{Self-Correction}
\begin{figure}[tbp]
  \centering
  \includegraphics[width=1.0\linewidth]{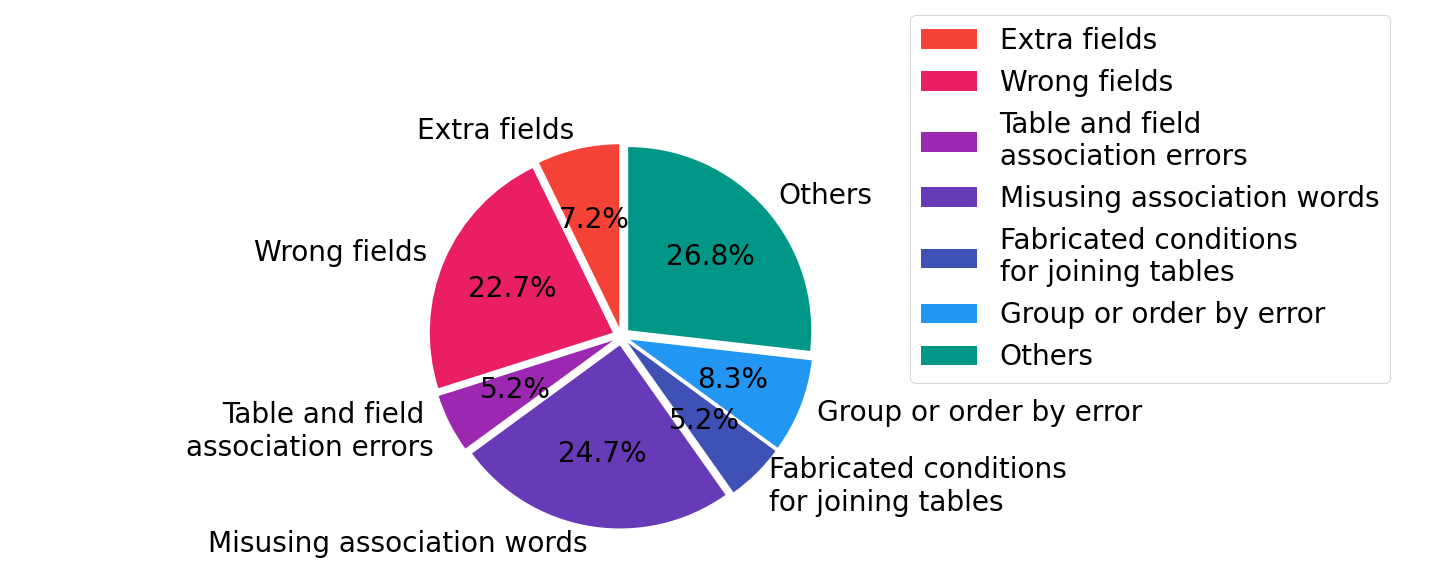}
  \caption{The statistics of SQL errors based on information determination, classification \& hint and SQL generation steps}
  % \Description{A figure shows the overall flow chart of the OlaGPT model.}
  \label{fig:sql_error}
\end{figure}

After analyzing the results generated by the LLM in the previous steps, it is found that although the decomposition in the previous steps reduces redundant information and focuses the effective attention of the LLM, resulting in an improvement in accuracy(The outcome closely aligns with the most recent baseline method, achieving 83.4\%. This can be observed in the 'w/0 active \& correct' result in Table \ref{tab:ablation_study}.), there are still some issues in field selection, table joining, aggregation, etc. 
The analysis of 98 incorrect questions reveals the proportion of error types, as depicted in Figure \ref{fig:sql_error}. Notably, several significant error points emerge: 1) Extra fields: The LLM often selects an excessive number of fields, rather than limiting its selection to those pertinent to the question; 2) Incorrect fields: For instance, when faced with fields bearing identical names across different tables, the alias may be omitted, leading to errors; 3) Table and field association errors: there may be inconsistencies between the tables and fields used; 4) Misuse of association words: For example, there is a tendency to habitually use 'left join' in place of 'join'; 5) Fabricated conditions for table joins; 6) Group or order by errors: Mistakes such as incorrect aggregation fields and conditions may be encountered.

In response to the aforementioned issues, we establish prompts as demonstrated in the Self-Correction step in Figure \ref{fig:sys}. Instead of using entirely generic prompts like DIN-SQL~\cite{pourreza2023dinsql}, we analyze the ability threshold of LLMs (i.e., the types of mistakes they frequently make) and set up relevant error prompts specifically targeting those issues, following the generic prompt points.

\subsection{Active Learning}
The model itself has a limited capability threshold and the circles it specializes in do not encompass all problem types in SQL as shown in the Active Learning module in Figure \ref{fig:sys}. In the experiment, we discover that the model is more prone to errors for certain problem types (e.g., extremum problems), often producing fixed-paradigm yet incorrect answers. 
We aim to determine whether the model's capability thresholds can be expanded through the Active Learning Module to better meet the requirements. 
The model's generalization ability is used to present the wrong case and the standard answer to the model to learn the error correction paradigm.

Specifically, we sample part of the training set for the preamble step of the test, analyze the error type of the LLM to organize and summarize some of the samples, and ultimately select three fixed typical samples as the error case. The case format $C$ follows $<Q, O, A>$, where $Q$ represents the question of one case, $O$ represents the origin SQL that is generated by the LLM and $A$ is the answer including whether it needs to modify and the modified SQL.
The final active learning prompt follows $<I, C, Q>$, where $I$ represents the instruction that cues the need for the model to accomplish this task, $C$ represents the three fixed error cases and $Q$ is the question that should be answered.

\begin{table*}[!ht]
  \caption{Comparison of previous methods and our method in terms of execution accuracy on Spider dataset.}
  \label{tab:overall_results}
  \centering
  \small
  \begin{tabular}{c|ccc|cc}
    \toprule
    Method & Zero-Shot & Few-Shot & Fine-tuning & Dev & Test \\
    \midrule
    T5-3B + PICARD \cite{scholak2021picard} &  &  & \checkmark & 79.3 & 75.1 \\
    Graphix-3B + PICARD \cite{li2023graphix} &  &  & \checkmark & 81.0 & 77.6 \\
    RESDSQL-3B + NatSQL \cite{li2023resdsql} &  &  & \checkmark & \underline{84.1} & 79.9 \\

    ChatGPT \cite{ouyang2022training} & \checkmark &  &  & 74.4 & - \\ 

    GPT-4 \cite{2023gpt} & \checkmark &  &  & 72.3 & - \\ 
    
    C3 + ChatGPT \cite{dong2023c3} & \checkmark &  &  & 81.2 & 82.3 \\ 
    DIN-SQL + GPT-4 \cite{pourreza2023dinsql} &  & \checkmark &  & 83.5 & 85.3 \\
    DAIL-SQL + GPT-4 \cite{gao2023text} & & \checkmark &  & 83.1 & 86.2 \\
    DAIL-SQL + GPT-4 + SC \cite{gao2023text} & & \checkmark &  & \underline{83.6} & \underline{86.6}\\
    % 0.802
    MAC-SQL \cite{wang2023mac} &  & \checkmark  &  & 78.6 & -\\
    \midrule
    % \ours + GPT-3.5 & & \checkmark &  & 80.4 & - \\
    \ours + GPT-4 & & \checkmark &  & \textbf{85.4} & \textbf{87.1} \\
    \bottomrule
  \end{tabular}
\end{table*}

\section{Experiment}
In this section, we evaluate the proposed algorithm for the text-to-SQL task.
Extensive experiments have been conducted to answer the following research questions: \textbf{ RQ1}. How does \ours perform vs. state-of-the-art baselines? \textbf{RQ2}. Whether each module of our method works effectively and how are they impacting the task? \xieyzh{\textbf{RQ3}. What is the token and time cost of the method?} \textbf{RQ4}. How do model parameters like the number of few-shots or the information filter layers affect the method?

\subsection{Experiment Setup}
\subsubsection{Implementation Details}
We use OpenAI ChatGPT-3.5 and GPT-4\footnote{gpt-3.5-turbo-0613 for ChatGPT and gpt4-0613 for GPT-4.} as our base models, which can be assessed through official API. For all GPT-based baselines, we utilize the same API to reproduce the models, ensuring fairness in comparison.
Our code is available at \url{https://github.com/FlyingFeather/DEA-SQL}.

\subsubsection{Datasets}
\paragraph{Spider~\cite{yu2018spider}.}
% In our experiments, we primarily utilize the \textbf{Spider} dataset~\cite{yu2018spider} to evaluate the performance of our methods. 
Spider is a large-scale dataset for complex, cross-domain semantic parsing and text-to-SQL tasks. It consists of 10,181 questions (8,659 examples in the training set and 1,034 examples in the development set ({Spider-dev}).) and 5,693 unique complex SQL queries, involving 200 databases and multiple tables covering 138 different domains. 
\paragraph{Spider Realistic~\cite{deng2021structure}.}
% In addition, we conduct experiments on \textbf{Spider-Realistic}~\cite{deng2021structure}, a more challenging version of the Spider-dev. 
Spider Realistic is a more challenging version of the Spider-dev. It modifies the natural language questions in Spider by removing or paraphrasing explicitly mentioned column names to generate a more realistic dataset reflecting real-world scenarios, where questions rarely include explicitly mentioned column names. The final dataset comprises a total of 508 question-query pairs.
\xieyzh{
\paragraph{Bird~\cite{li2024can}.} Bird is a comprehensive dataset, containing 12,751 distinct question-SQL pairings, spread across 95 large-scale databases that collectively amount to 33.4 GB in size. This dataset covers an expansive range of over 37 professional fields such as blockchain, hockey, healthcare, and education. In a bid to help models generate precise SQL queries, Bird integrates external knowledge as an auxiliary resource. This external knowledge is specifically derived from four sources: numerical reasoning, domain-specific knowledge, synonyms, and value illustration.
}

\subsubsection{Evaluation}
There are two commonly used evaluation metrics for the text-to-SQL task: (1) Exact Match Accuracy (EM). This metric requires each component of the predicted SQL to be identical to the corresponding component of the gold SQL, disregarding the specific values in the query. (2) Execution Accuracy (EX). EX assesses the correctness of the execution result of the predicted SQL, typically offering a more precise evaluation than EM. Given that there might be multiple valid SQL queries with different writing styles for a single given question, EM might not effectively reflect the efficacy of the predicted SQL. As a result, we exclusively employ EX as our primary evaluation metric. Following previous work~\cite{ruiqi20}, we utilize the evaluation scripts available at \url{https://github.com/taoyds/test-suite-sql-eval}.

% \subsubsection{Hyperparameter}

\subsection{Perfermance (RQ1)}
To validate the effectiveness of \ours in text-to-SQL tasks, a comprehensive comparison was performed on the Spider, Spider-Realistic, and Bird datasets. We performed most of our experiments on the dev dataset of the Spider dataset, and the experiment was completed just coinciding with the opening of the test set.
The experimental results are summarized in Table \ref{tab:overall_results}, \ref{tab:overall_results_sr} and \ref{tab:overall_results_bird}\footnote{Due to cost constraints, we only provide the results of DEA-SQL+ChatGPT+zeroshot-SC(reduce Self-Correction module) on Bird Dev dataset}, where the best performance is shown in boldface. Based on the experimental results, we have several findings.

In terms of execution accuracy in the Spider dataset, our approach based on workflow outperforms the existing baselines as shown in Table \ref{tab:overall_results}), and the approach achieves a nice boost over the direct use of prompt on different pedestal models (Llama, WizardCoder, and CodeLlama) as shown in Table \ref{tab:diff_base_model}). The effectiveness of the workflow paradigm based on decomposing for enhancing attention is effectively demonstrated.

\begin{table}[!ht]
    \caption{Comparison of previous methods and our method in terms of execution accuracy on Spider-Realistic dataset.}
    \label{tab:overall_results_sr}
    \centering
    \small
    \footnotesize
    \begin{tabular}{c|ccc|cc}
      \toprule
      % Method & Spider-Realistic \\
    Method & EX \\
      \midrule
      % T5-3B + PICARD  &  0 \\
      % Graphix-3B + PICARD    &  0 \\
      % RESDSQL-3B + NatSQL     &  81.9 \\
      % ChatGPT       &  0 \\ 
      % GPT-4                        &  0 \\ 
      C3 + ChatGPT          &  75.4 \\ 
      DIN-SQL + GPT-4       &  78.1 \\
      DAIL-SQL + GPT-4       &  75.6 \\
      DAIL-SQL + GPT-4 + SC    &  75.2 \\
      \midrule
      \ours + GPT-4                                  & \textbf{81.5} \\
      \bottomrule
    \end{tabular}
  \end{table}

Compared with traditional training models, our method also has good improvements, further proving the potential of LLMs in text-to-SQL tasks. Providing a good workflow paradigm based on decomposing for enhancing attention can effectively improve LLMs' performance in complex tasks.

In the Spider-Realistic dataset, which is more adapted to real-world scenarios and has more difficult question formulations, our approach is more stable and achieves better performance than other solutions based on LLMs. This validates the effectiveness of the two-stage information determination we proposed, which can mitigate the impact of different question formulations to some extent. 
The performance of the DAIL-SQL declines significantly on this dataset, indicating that the approach of relying solely on prompt optimization and few-shot retrieval has its limitations, making it more challenging to establish a good few-shot library and retrieval strategy for real-world problems.

\xieyzh{
On the experimental results of the more complex Bird dataset in Table \ref{tab:overall_results_bird}, our method still achieved good results, demonstrating considerable scalability.
Compared with directly using ChatGPT, we have increased by 15.13 percentage points; compared with MAC-SQL+ChatGPT, we have increased by 1.79 percentage points; even compared with DIN-SQL+GPT4, we have increased by 1.63 percentage points. We firmly believe that applying the method directly to GPT4 will yield greater improvements. Because of the multi-step method, when the accuracy of each step improves, the final accuracy improvement is cumulative.}

  \begin{table}[!ht]
    \caption{Comparison of previous methods and our method in terms of execution accuracy on the Bird Dev dataset.}
    \label{tab:overall_results_bird}
    \centering
    \small
    \footnotesize
    \begin{tabular}{c|ccc|cc}
      \toprule
    Method &  EX\\
      \midrule
      ChatGPT         & 37.22\\ 
      GPT4                 & 46.15\\
      DIN-SQL + GPT-4        & 50.72\\
      MAC-SQL + GPT-4 & 50.56 \\
      \midrule
      \begin{tabular}[c]{@{}c@{}}\ours + ChatGPT+ zeroshot-SC\\(reduce Self-Correction module)\end{tabular}   & \textbf{52.35} \\
      % \ours + ChatGPT + zeroshot(reduce Self-Correction module)   & 52.35 \\
      \bottomrule
    \end{tabular}
  \end{table}

We also investigated the improvement at different difficulty levels, as shown in Table \ref{tab:level}, there is a noticeable improvement at the hard and extra hard difficulty levels. This is partly due to the information determination module reducing irrelevant information interference, the problem classification module classifying and solving problems of different difficulty levels, and the self-correction and active learning step further expanding the ability threshold of LLM to solve problems.

\subsection{Ablation Study (RQ2)}
In that subsection, we ablate each module and evaluate the effectiveness of each module. As shown in Table \ref{tab:ablation_study}, the exclusion of either module makes the overall performance of the method degrade.

\begin{table}[!ht]
    \caption{Ablation study on Spider Dev dataset.}
    \label{tab:ablation_study}
    \centering
    % \small
    % \footnotesize
    \scriptsize
    \begin{tabular}{c|c|c|c|c|c}
      \toprule
      Method &  Easy & Medium & Hard & Extra & All \\
      \midrule
      \ours          & 0.887 & 0.895 & 0.856 & 0.705 & 0.856 \\ 
      w/o information filter         & 0.899 & 0.874 & 0.787 & 0.633 & 0.827 \\
      w/o classification & 0.875 & 0.904 & 0.805 & 0.639 & 0.838  \\
      w/o self-correct   & 0.895 & 0.881 & 0.839 & 0.663 &  0.842 \\
      w/o active learning & 0.903 & 0.892 & 0.822 & 0.663 & 0.846  \\
      w/o active \& correct  & 0.903 & 0.879 & 0.805 & 0.639 & 0.834  \\
      w/o H-FK & 0.891 & 0.886 & 0.851 & 0.651 & 0.843 \\
      % \hline
      \bottomrule
    \end{tabular}
  \end{table}

\begin{table}[!ht]
    \caption{Performance compared to our method different query across difficulty levels on Spider Dev dataset.}
    \label{tab:level}
    \centering
    % \small
    \scriptsize    
    \begin{tabular}{c|c|c|c|c|c}
      \toprule
      Method &  Easy & Medium & Hard & Extra & All \\
      \midrule
      C3-ChatGPT & 0.919 & 0.841 & 0.782 & 0.608 & 0.812\\
      DIN-SQL & 0.907 & 0.897 & 0.793 & 0.675 & 0.835\\ 
      DAIL-SQL & 0.907 & 0.897 & 0.753 & 0.62 & 0.831 \\ 
     \ours          & 0.891 & 0.892 & 0.845 & 0.705 & 0.854 \\ 
     % \ours          & 0.887 & 0.895 & 0.856 & 0.705 & 0.856 \\ 
     
      % \hline
      \bottomrule
    \end{tabular}
  \end{table}
  
\paragraph{The effect of sub-models.}
1) In \textbf{information determination}, the strategy of reducing irrelevant information has a positive effect on all types of questions except those of simple difficulty. It is particularly effective in improving the accuracy of hard and extra hard difficulty questions. By reducing irrelevant information to focus the attention of LLMs, we can effectively enhance performance in complex tasks.
2) The ablation method of the \textbf{classification} step is to remove the classification module while switching the hint and specification in the prompt structure to a fixed hint.
It is found that besides damaging the problem-solving ability of medium-difficulty problems, this step has positive effects on other problems, and the effects are significant in difficult and extremely difficult problems.
It is indicated that distinguishing problem types, using simple hints for simple problems and targeting complex hints for complex problems, can effectively improve the performance of the LLM.
3) The \textbf{active learning} and \textbf{self-correction} modules had a negative effect on easy problems, but a positive effect on hard and extra hard problems. 
This is also expected, as these two modules are designed to increase the capability threshold of the original base model, but may to some extent impair the ability to solve other easy questions.
4) Adding \textbf{H-FK} prompt can effectively improve the performance of complex difficulty, which reduces the difficulty of finding the right complex concatenated table condition to some extent for the model.
To summarize, for easy questions there is no need for overly complex processes and prompts, while for relatively complex questions, the multi-step workflow approach of narrowing down the information to enhance the model's attention works well.

\paragraph{The effect of different few-shot scheme.}
To explore the effectiveness of few-shot retrieval methods, we conducted experimental comparisons between different retrieval schemes and zero-shot. In the experiments, it was found that different few-shots have a significant impact on the final accuracy of the results, and choosing poor examples may lead to adverse effects. We explored four retrieval schemes: random, question similarity(ques\_sim), template similarity(tem\_sim), and template similarity without classification (tem\_sim\_wo) to find more suitable samples. As shown in Table \ref{tab:few-shot_mode}, random is a little detrimental to overall performance, while the retrieval strategy based on question template similarity in the combined question classification retrieval library yields the best results.
Question templates also essentially provide a simple classification for the questions, and relying on question classification methods makes it easier to find the most relevant questions and their solutions, thereby stimulating the capabilities of LLMs.

\xieyzh{
\subsection{Cost Analysis (RQ3)}
\begin{table}[!ht]
    \caption{The token and time cost analysis. Inference Time is the average time of running the same case three times using ChatGPT. \xieyzh{\nours-C represents the removal of the classification module in the \nours method.} }
    \label{tab:cost_analysis}
    \centering
    \scriptsize
    \begin{tabular}{c|c|c|c}
      \toprule
      Method  & Avg. Token Num &  Inference Time &  EX  \\
      \midrule
      C3    & \underline{~2803}  & 19.34s & 81.2 \\
      DIN-SQL & ~9126 & \underline{4.37s} & 83.5 \\
      DAIL       & \textbf{~700}  &  - &  83.6 \\
      MAC-SQL & ~4043   & 5.45s  & 78.6 \\
      \midrule
      \ours &  ~5611 & 4.55s & \textbf{85.4} \\
       \nours-C &  ~3324 & \textbf{4.16s} & \underline{83.8} \\
      % \hline
      \bottomrule
    \end{tabular}
  \end{table}
Indeed, the cost is a factor that should be considered in practical use, and we conduct the experiment for an analysis of token and time costs to substantiate the usability of the framework we propose. It can be clearly seen from the Table \ref{tab:cost_analysis}:
1) Our method consumes less in terms of inference time, ensuring the efficiency of our method in real applications.
2) In terms of token consumption, we need some examples for the model to learn from the context, which leads to some increase in tokens. However, as can be seen, it is still more economical than DIN-SQL and brings the best model effect. We believe this is an acceptable trade-off (in addition, the cost and effect balance can be controlled by reducing steps or using traditional solutions to replace some steps).
Our core point is the attention-focused workflow paradigm in in-context learning. In practical use scenarios, we can fully decide to replace some modules with other solutions to reduce overall consumption. 
}

\subsection{Parameters Analysis (RQ4)} 
\begin{figure}[htbp]
  \centering
  \begin{minipage}[t]{0.4\linewidth}
    \centering
    \subfigure[The number of few-shot\label{fig:sql_fewshot_nums}]{
      \includegraphics[width=0.9\linewidth]{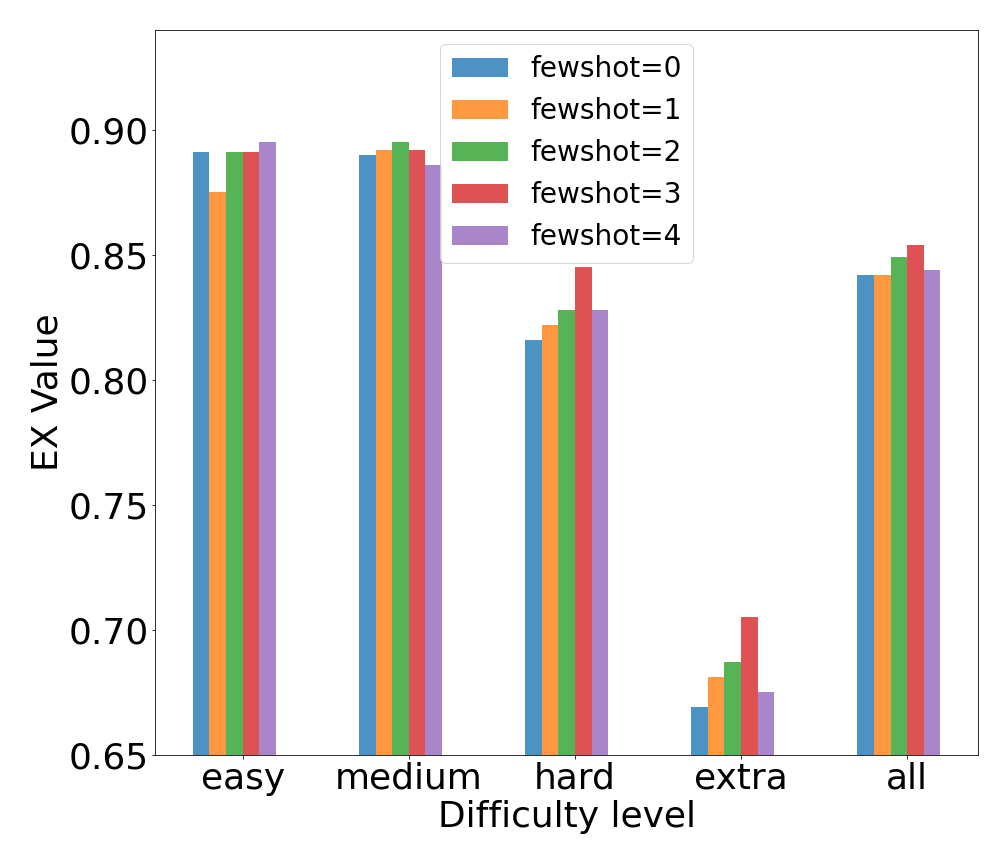}
    }
  \end{minipage}
  \begin{minipage}[t]{0.4\linewidth}
    \centering
    \subfigure[The number of information filter layer\label{fig:sql_fc}]{
      \includegraphics[width=0.9\linewidth]{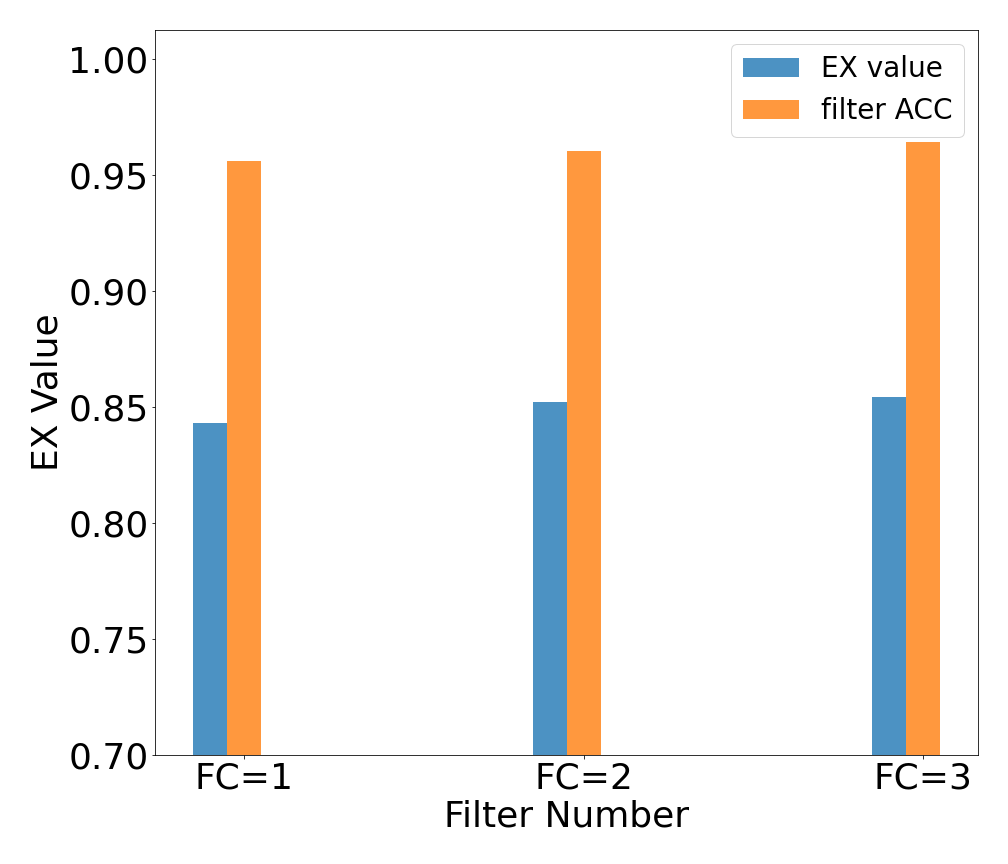}
    }
  \end{minipage}
  % fewshot的探索
  \caption{Parameters analysis results.}
  \label{fig:two_images}
\end{figure}

In the parameters experiment, we investigated the number of few-shots for the SQL generation step, and the results in Figure \ref{fig:sql_fewshot_nums} found that the performance is better when \#fewshot=3. This indicates that within a certain range, as the number of effective samples increases, the accuracy of the model improves.
The accuracy of the information filter also greatly affects the final accuracy rate. The current method based on LLM has not yet been able to achieve 100\% accuracy. It can be seen that with the improvement of the accuracy of the information filter, the final accuracy rate also has a corresponding improvement as shown in Figure \ref{fig:sql_fc}.

\begin{table}[!ht]
    \caption{The few-shot mode study on Spider Dev dataset, where the number of few-shots is 3.}
    \label{tab:few-shot_mode}
    \centering
    \scriptsize
    \begin{tabular}{c|c|c|c|c|c}
      \toprule
      Method  & Easy & Medium & Hard & Extra & All \\
      \midrule
      
      zero\_shot    & 0.891  & 0.890 & 0.816 & 0.669 & 0.842\\
      random & 0.879 & 0.890 & 0.862 & 0.693 & 0.851 \\
      ques\_sim       & 0.879  & 0.901 & 0.833 & 0.645 & 0.843 \\
      tem\_sim &  0.891  & 0.892 & 0.845 & 0.705 & 0.854 \\
       tem\_sim\_wo & 0.895 & 0.888 & 0.822 & 0.657 & 0.841 \\
      % \hline
      \bottomrule
    \end{tabular}
  \end{table}

\section{Related Work}

\paragraph{Text-to-SQL Parsing}
Text-to-SQL is a semantic parsing task that translates users' natural language questions into appropriate SQL queries.
This allows ubiquitous relational data to be accessed by a broader range of non-technical users. In the past few years, inspired by the success of deep learning, research in text-to-SQL has primarily focused on constructing well-designed deep neural networks~\citep{guo2019towards, wang2020rat, cao-etal-2021-lgesql, scholak2021picard, li2023resdsql}. 
RATSQL~\citep{wang2020rat} defines a question-schema graph and employs a relation-aware self-attention mechanism during the encoding process to jointly learn representations of the question words, schema items, and edge relationships. PICARD~\citep{scholak2021picard} is a sequence-to-sequence model. The PICARD model rejects invalid tokens at each decoding step and constrains the generated results within a certain output space, thereby reducing the number of invalid SQL queries. 
% RESDSQL~\citep{li2023resdsql} reduces the complexity of text-to-SQL by implicitly constraining SQL parsing through schema linking and skeleton parsing techniques.
Although these supervised learning neural models have achieved impressive performance on the text-to-SQL task, they are typically trained on a large training set and evaluated on test examples. However, acquiring annotated text-to-SQL data is costly. Additionally, training and fine-tuning the models entail significant engineering efforts and consume substantial computational resources.

\paragraph{In-context Learning for Text-to-SQL}
With the widespread adoption of LLMs, recent work has explored in context learning for text-to-SQL from various perspectives~\citep{gao2023text, liu2023comprehensive, chang2023selective}. 
In Context learning enables pre-trained LLMs to perform text-to-SQL without relying on supervised samples by providing zero or a few training examples (NLQ-SQL pairs) as demonstrations. In this domain, 
~\citet{rajkumar2022evaluating} and  ~\citet{chang2023prompt} investigate effective methods for representing the database within prompts in context learning.
~\citet{pourreza2023dinsql} and ~\citet{ wang2023mac} improve the ability of text-to-SQL parsing through intermediate reasoning steps.
% ~\citet{chen2023teaching} and ~\citet{pourreza2023dinsql} explore the self-debugging ability of LLMs to correct potentially erroneous SQL statements. 
Other studies~\citep{tai2023exploring, zhang2023act} focus on enhancing the semantic understanding capabilities of LLMs using chain-of-thought techniques, thereby addressing text-to-SQL problems more effectively.
However, these studies tend to focus more on activating the general capabilities of LLMs, lacking a comprehensive consideration of SQL tasks. Additionally, the presence of irrelevant tables and fields in the database schema can interfere with the generation of correct SQL queries. To address these issues, we propose a complete SQL generation workflow based on prior knowledge of attention-focusing and irrelevant information filtering. 
\section{Conclusion}

Prompting techniques have unleashed the potential of LLMs, with most past research focusing on improving single-step prompting. In the NLP field, structured languages like SQL pose a greater challenge than plain text. We introduce the workflow prompting paradigm to enhance the performance of LLMs in text-to-SQL tasks. By decomposing steps, reducing irrelevant information, and transforming the complex problem into a classification task, we can focus the attention of LLMs. Self-correction and active learning modules are designed according to the capability thresholds of LLMs, with different sub-modules performing their respective roles, further improving the upper limit of LLM-based prompting methods in text-to-SQL tasks. Adequate experiments on Spider, Spider-Realistic, and Bird datasets demonstrate that our approach significantly improves prompting performance, outperforming state-of-the-art fine-tuning and LLM-based methods. This illustrates the effectiveness of the workflow prompting paradigm in enhancing the problem-solving scope of LLMs.

\clearpage

\section*{Limitations}

1. \textbf{Cost Issues}: Our proposed method adopts a decomposition workflow as the backbone, and when using GPT-4 to respond to natural language questions, there might be some differences in cost and latency compared to the zero-shot approach, thus a trade-off between performance and cost is necessary.
\\
2. \textbf{Instability}: Another limitation of text-to-SQL methods based on LLMs, is their inherent instability due to the probabilistic nature of their generation process. As these models rely on probability distributions to generate output, they can sometimes produce inconsistent or unpredictable results. Considering more deterministic SQL statement generation can be conducted in future work.

% \clearpage
\bibliography{ref}

\appendix
\onecolumn

\section{Experimental Supplement}
\subsection{Baselines}
We select conventionally and recently published text-to-SQL baselines for model comparison, which can be briefly categorized into two groups: (1) training model-based programs~\cite{scholak2021picard,li2023graphix,li2023resdsql} and (2) prompt engineering-based programs~\cite{dong2023c3,pourreza2023dinsql,gao2023text,wang2023mac}.

\begin{itemize}
% [leftmargin=*]
\item {\textbf{T5-3B + PICARD }} \cite{scholak2021picard}
PICARD is a sequence-to-sequence model. It rejects invalid tokens at each decoding step and constrains the generated results within a certain output space, thereby reducing the number of invalid SQL queries. 
\item {\textbf{Graphix-3B + PICARD }} \cite{li2023graphix} Based on the pre-trained T5 model, Graphix is designed to encode the combination of semantic and structural information via graph neural networks.
\item {\textbf{RESDSQL-3B + NatSQL }} \cite{li2023resdsql}
RESDSQL reduces the complexity of text-to-SQL by implicitly constraining SQL parsing through schema linking and skeleton parsing techniques. It is the best method based on fine-tuning.
\item {\textbf{ChatGPT }} \cite{ouyang2022training}
We use ChatGPT in a simple zero-shot paradigm to evaluate its capability on the text-to-SQL task.
\item {\textbf{GPT-4 }} \cite{2023gpt}
The base model is changed to GPT-4 in a simple zero-shot paradigm to evaluate the capability on the text-to-SQL task.
\item {\textbf{C3 + ChatGPT }} C3~\cite{dong2023c3} proposes a ChatGPT-based zero-shot Text-to-SQL method with three key components: Clear Prompting (CP), Calibration with Hints (CH), and Consistent Output (CO).
% 提出三个关键部分：清晰提示(CP)、提示校准(CH)和一致输出(CO)
\item {\textbf{DIN-SQL + GPT-4}} DIN-SQL~\cite{pourreza2023dinsql} utilizes multi-step decomposition and self-correction to effectively improve the performance of context-based learning methods on text-to-SQL tasks.
\item {\textbf{DAIL-SQL + GPT-4 + SC }} DAIL-SQL~\cite{gao2023text} systematically and extensively compares existing prompting strategies and proposes a novel integrated solution. It is the best method based on LLMs.
\item {\textbf{MAC-SQL }} MAC-SQL~\cite{wang2023mac} proposes an innovative LLM-based multi-agent framework for the Text-to-SQL task. It consists of three agents: the Selector, which condenses databases and retains relevant table schemas; the Decomposer, breaking down complex questions into simpler sub-problems; and the Refiner, responsible for validating and refining SQL.
\end{itemize}

\subsection{Intergrated with other LLMs}

In the main article, to better explore the performance of LLMs, we chose the advanced GPT4 model as the base model for experimentation, while this section mainly supplements the introduction of experimental results using other models as the base model. As can be seen from Table \ref{tab:diff_base_model}, the use of our proposed \ours method effectively improved the model's execution accuracy in the text-to-SQL task, proving the effectiveness of the decomposing for enhanced attention workflow paradigm.

    \begin{table*}[!ht]
        \centering
        \caption{Performance comparison of different LLMs when applied \nours.}
        \label{tab:diff_base_model}
        \begin{tabular}{cccccccc}
          \toprule
          LLM & \textbf{Easy} & \textbf{Medium} & \textbf{Hard} & \textbf{Extra} & \textbf{All} \\
          % \cmidrule(lr){2-4}\cmidrule(lr){4-5}
          % & EX & EM & EX & EM \\
          \midrule
          Llama-2-13B-chat       & 0.762     & 0.500     & 0.287     & \textbf{0.157}    & 0.472 \\
        \textbf{+\nours}     & \textbf{0.823}     & \textbf{0.585}     & \textbf{0.374}     & 0.145 & \textbf{0.536} \\
          \midrule
          WizardCoder-15B-V1.0       & 0.859     & 0.774     & 0.540     & 0.289   & 0.677  \\
          \textbf{+\nours}     & \textbf{0.879}     & \textbf{0.816}     & \textbf{0.557}     & \textbf{0.319} & \textbf{0.708} \\
          \midrule
          CodeLlama-13B-Instruct       & 0.806     & 0.749     & 0.529     & 0.380  & 0.666   \\
        +\textbf{\nours}     & \textbf{0.871}     & \textbf{0.796}     & \textbf{0.546}     & \textbf{0.380} & \textbf{0.705} \\
          \bottomrule
        \end{tabular}
      \end{table*}

% todo
% codellama. llama2, wizardcoder

\section{Prompts}
% 为了便于更能理解我们的方法，本节全面列出brarinSQL整个工作流方法中每个步骤流的提示信息。
In order to facilitate a better understanding of our approach, this section provides a comprehensive list of prompts for each of the steps in \nours's entire workflow approach.

\subsubsection{Identify Problem Elements Prompt of Spider Dataset}

Assuming that you are a natural language processing expert and statistician, and a data analyst, please understand the business requirements and break down the requirements description into statistical elements. It is required to break down user problems into entities, and the main information in the original problem cannot be lost. \\
\\
\#\#\# Here are some examples: \\
What is the name of the staff that is in charge of the attraction named 'US museum'? \\
output: \{"entities": ["staff", "the attraction named 'US museum'"], "query": "the name of the staff that is in charge of the attraction named \"US museum\""\} \\
\\
How many heads of the departments are older than 56 ? \\
output: \{"entities": ["age older than 56", "number of heads of the departments"], "query": "Number of department heads over 56 years old"\} \\
\\
List the name, born state and age of the heads of departments ordered by age.\\
output: \{"entities": ["name of the heads of departments", "born state of the heads of departments", "age of the heads of departments", "age"], "query": "List the name, born state and age of the heads of departments ordered by age."\}\\
\\
what is the average, minimum, and maximum age of all singers from Chinese?\\
output: \{"entities": ["Chinese", "age of all singers"], "query": "The average, minimum, and maximum age of all singers from Chinese"\}\\
\\
Return the different descriptions of formulas that has been used in the textbook.\\
output: \{"entities": ["the different descriptions of formulas", "formulas", "used in the textbook"], "query": "The different descriptions of formulas that has been used in the textbook"\}\\
\\
What are the details of the markets that can be accessed by walk or bus?\\
output: \{"entities": ["the details of the markets", "can be accessed by walk or busk"], "query": "The details of the markets that can be accessed by walk or bus"\}\\
\\
Show the name of colleges that have at least two players.\\
output: \{"entities": ["the name of colleges", "players"], "query": "The name of colleges that have at least two players"\}\\
\\
How many gold medals has the club with the most coaches won?\\
output: \{"entities": ["gold medals", "club", "coaches"], "query": "The number of gold medals has the club with the most coaches won"\}\\
\\
List the nominees that have been nominated more than two musicals.\\
output: \{"entities": ["nominees", "nominees that have been nominated", "musicals"], "query": "The nominees that have been nominated more than two musicals"\}\\
\\
\#\#\# Please be sure to follow the following specifications:\\
1."entities" refers to all entities in the requirements,  including all description information in the requirements.\\
2.Your output must be output in json format, and only this json needs to be returned. It needs to include all fields in json. The json format is as follows: 
\{"entities":[entities], "query":"Rewritten question, removing unnecessary content"\}\\
\\
\textbf{[query input]}\\
output:
\\

\subsubsection{Identify Problem Elements Prompt of Bird Dataset}

Assuming that you are a natural language processing expert and statistician, and a data analyst, please understand the business requirements and break down the requirements description into statistical elements. It is required to break down user problems into entities, and the main information in the original problem cannot be lost.\\
\\
\#\#\# Here are some examples:\\
query: Which year has the least number of movies that was released and what is the title of the movie in that year that has the highest number of rating score of 1?\\
hint: least number of movies refers to MIN(movie\_release\_year); highest rating score refers to MAX(SUM(movie\_id) where rating\_score = '1')\\
output: {{"query\_entities": ["year", "number of movies that was released", "the title of the movie", "rating score"], "hint\_entities": ["movie\_release\_year", "movie\_id", "rating\_score"], "query": "The year with the least number of movies released is identified, and the title of the movie with the highest rating score of 1 in that year is provided."}}\\
\\
query: What's the description of user 85981819's movie list with the most followers?\\
hint: user 85981819 refers to user\_id = 85981819; most followers refers to Max(list\_followers); description refers to list\_descriptions;\\
output: {{"query\_entities": ["the description of user", "movie list", "followers"], "hint\_entities": ["user\_id", "list\_followers", "list\_descriptions"], "query": "The description of user 85981819's movie list with the most followers is requested."}}\\
\\
query: Who is the youngest player to have won the Purple Cap?\\
hint: Who refers to Player\_Name; youngest player to have won the Purple Cap refers to min(subtract(Season\_Year, DOB))\\
output: {{"query\_entities": ["player", "the Purple Cap"], "hint\_entities": ["Player\_Name", "Season\_Year", "DOB"], "query": "The name of youngest player to have won the Purple Cap."}}\\
\\
How old is Ishan Kishan in 2022?\\
hint: old refers to SUBTRACT(2022, SUBSTR(DOB, 1, 4)); Ishan Kishan refers to Player\_Name = 'Ishan Kishan';\\
output: {{"query\_entities": ["age", "Ishan Kishan", "2022"], "hint\_entities": ["DOB", "Player\_Name"], "query": "The age of Ishan Kishan in 2022"}}\\
\\
query: What's the average rating score of the movie "A Way of Life"?\\
hint: A Way of Life' is movie\_title; average rating score = Divide (Sum(rating\_score), Count(rating\_id));\\
output: {{"query\_entities": ["rating score", "movie", "A Way of Life"], "hint\_entities": ["movie\_title", "rating\_score", "rating\_id"], "query": "The average rating score of the movie \"A Way of Life\""}}\\
\\
\#\#\# Please be sure to follow the following specifications:\\
1."query\_entities" refers to all entities in the query, including all description information in the query.\\
2."hint\_entities" refers to all entities in the hint, including all description information in the hint.\\
3.Your output must be output in json format, and only this json needs to be returned. It needs to include all fields in json. The json format is as follows: \\
{{"query\_entities":[query entities], "hint\_entities":[hint entities], "query":"Rewritten question, removing unnecessary content"}}\\
\\
query: \textbf{[query input]}\\
hint: \textbf{[hint]}\\
output:

\subsection{Information Filter Prompt}
\textbf{[table\_info]} \\
User question: \textbf{[query]} \\
Entity information: [\textbf{limitation]} \\
\#\#\# need \\
You are a data analyst. In business, you need to use the above table information to complete a SQL query code to solve user problems. I would like to ask you to first match the table fields or calculation methods required by the \textbf{[limitation]} entity, and then determine the calculation method of \textbf{[main\_metric]}, and finally determine the required table and all related field information and give some key information for writing SQL. \\
Note that all table names must be their original names, and the output of field names must be the original field names in the table. \\
\#\#\# Please be sure to comply with the following specifications\\
1. Element matching needs to output the most related table fields (one or more) or calculation methods and required field names required by the entities; yyy1 is the table field that needs to be selected to calculate the entity and the answer is in the form of ```colunm\_name```. Note that an entity may require multiple fields;\\
2. bbb is the calculation method of \textbf{[main\_metric]};\\
3. Required table information: Not all tables may need to be selected, depending on the specific problem.\\
3.1. Select the table and related fields based on the user questions, entity information and element matching information you have given above;\\
3.2. The where statement condition only gives the conditions of the corresponding table;\\
3.3. All field names required by SQL under the table must include the fields actually needed under the corresponding table. Note that you cannot select fields that are not under the previous table name, and do not select all fields. You must include all the fields that are needed for the table;\\
4. Multi-table joint fields and conditions need to find out the associated fields and conditions between multiple tables from the above table information;\\
5. "All fields" must to include all the fields actually used in sql !!! You must include all the fields that are needed for the table;\\
6. Think step by step, and finally summarize that your output is only in the given json format: {"Element matching": {output\_format}, "{main\_metric} calculation method": "bbb", "Required table information": [{"Table name": "xxx", "where statement condition": "ccc", "All field names required by SQL under this table": ["yyy1", "yyy2", "yyy3"]}, {"Table name": "xxx", "where statement condition": "ccc", "All field names required by SQL under this table": ["yyy1", "yyy2", "yyy3"]}], "Multiple table joint fields and conditions": "ccc", "sql": "ddd", "All fields": ["yyy1", "yyy2", "yyy3"]} \\

\subsection{Question Classification Prompt}
For the given question that requires writing SQL, classify it with two labels. You can choose the first label from NON-JOIN and JOIN and choose the second label from NON-NESTED and NESTED.\\
\#\#\# Some table infos and examples\\
Q: What are the names and revenues of the companies with the highest revenues in each headquarter city?\\
table\_info: CREATE TABLE MANUFACTURERS (\\
code INTEGER\\
name VARCHAR(255) NOT NULL\\
headquarter VARCHAR(255) NOT NULL\\
founder VARCHAR(255) NOT NULL\\
revenue REAL\\
PRIMARY KEY (code)   \\
);\\
\\
CREATE TABLE PRODUCTS (\\
code INTEGER\\
name VARCHAR(255) NOT NULL \\
price DECIMAL NOT NULL \\
manufacturer INTEGER NOT NULL\\
PRIMARY KEY (code), \\
FOREIGN KEY (manufacturer) REFERENCES Manufacturers(code)\\
);\\
A: Let’s think step by step. The SQL query for the question 'What are the names and revenues of the companies with the highest revenues in each headquarter city?' needs these tables and columns = [MANUFACTURERS.name, MANUFACTURERS.revenue, MANUFACTURERS.headquarter], so we don’t need joint condition and label it as NON-JOIN. Plus, it doesn’t require nested queries with (INTERSECT, UNION, EXCEPT, IN, NOT IN). so we label it as NON-NESTED.\\
Thus the SQL query can be classified as NON-JOIN, NON-NESTED\\
Label: NON-JOIN, NON-NESTED\\
\\
Q: Which studios have an average gross of over 4500000?\\
table\_info: CREATE TABLE FILM (\\
studio text\\
gross\_in\_dollar int\\
PRIMARY KEY (Film\_ID)\\
);\\
A: Let’s think step by step. The SQL query for the question 'Which studios have an average gross of over 4500000?' needs these table and column = [FILM.studio, AVG(FILM.gross\_in\_dollar)], so we don’t need joint condition and label it as NON-JOIN. Plus, it doesn’t require nested queries with (INTERSECT, UNION, EXCEPT, IN, NOT IN). So we label it as NON-NESTED.\\
Thus the SQL query can be classified as NON-JOIN, NON-NESTED\\
Label: NON-JOIN, NON-NESTED\\
\\
Q: What are the products with the maximum page size A4 that also have a pages per minute color smaller than 5?\\
table\_info: CREATE TABLE PRODUCT (\\
product\_id int\\
product text\\
dimensions text\\
dpi real\\
pages\_per\_minute\_color real\\
max\_page\_size text\\
interface text\\
PRIMARY KEY (product\_id)\\
);\\
A: Let’s think step by step. The SQL query for the question 'What are the products with the maximum page size A4 that also have a pages per minute color smaller than 5?' needs these table and columns = [PRODUCT.product, PRODUCT.max\_page\_size, PRODUCT.pages\_per\_minute\_color], so we don’t need joint condition and label it as NON-JOIN. Plus, it requires nested queries with (INTERSECT, UNION, EXCEPT, IN, NOT IN), and we need the answer to the questions = ['What is the maximum page size A4 of the products']. So it need nested queries and we label it as NESTED.\\
Thus the SQL query can be classified as NON-JOIN, NESTED.\\
Label: NON-JOIN, NESTED\\
\\
Q: Show names for all stadiums except for stadiums having a concert in year 2014.\\
table\_info: CREATE TABLE STADIUM (\\
stadium\_ID int\\
location text\\
name text\\
capacity int\\
highest int\\
lowest int\\
average int\\
PRIMARY KEY (Stadium\_ID)\\
);\\
\\
CREATE TABLE CONCERT (\\
concert\_ID int\\
concert\_Name text\\
theme text\\
stadium\_ID text\\
year text\\
PRIMARY KEY (concert\_ID),\\
FOREIGN KEY (stadium\_ID) REFERENCES stadium(stadium\_ID)\\
);\\
A: Let’s think step by step. The SQL query for the question 'Show names for all stadiums except for stadiums having a concert in year 2014.' needs these table and columns = [STADIUM.name, CONCERT.year], so we need a JOIN operation on the STADIUM and CONCERT tables using the stadium\_ID column because we we need to exclude stadiums with concerts in 2014. So we label it as JOIN. Plus, it requires nested queries with (INTERSECT, UNION, EXCEPT, IN, NOT IN), and we need the answer to the questions = ['What is the stadiums having a concert in year 2014']. So it need nested queries and we label it as NESTED.\\
Thus the SQL query can be classified as JOIN, NESTED.\\
Label: JOIN, NESTED\\
\\
Q: How many songs have a shared vocal?\\
table\_info: CREATE TABLE SONGS ( \\
SongId INTEGER PRIMARY KEY, \\
);\\
\\
CREATE TABLE VOCALS ( \\
SongId INTEGER \\
Bandmate INTEGER \\
PRIMARY KEY(SongId, Bandmate),\\
FOREIGN KEY (SongId) REFERENCES Songs(SongId),\\
FOREIGN KEY (Bandmate) REFERENCES Band(Id)\\
);\\
A: Let’s think step by step. The SQL query for the question 'How many songs have a shared vocal?' needs these table and columns = [SONGS.SongId, VOCALS.Bandmate], so we need a JOIN operation on the SONGS and VOCALS tables using the SongId column because we need to count the number of songs with a shared vocal. So we label it as JOIN. Plus, it does not require nested queries with (INTERSECT, UNION, EXCEPT, IN, NOT IN), so we label it as NON-NESTED.\\
Thus the SQL query can be classified as JOIN, NON-NESTED.\\
Label: JOIN, NON-NESTED\\
\\
Q: How many users who did not leave any review.\\
table\_info: CREATE TABLE USERACCT (\\
u\_id integer NOT NULL\\
name varchar(128) DEFAULT NULL\\
PRIMARY KEY (u\_id)\\
);\\
\\
CREATE TABLE REVIEW (\\
a\_id integer NOT NULL PRIMARY KEY\\
u\_id integer NOT NULL\\
FOREIGN KEY (u\_id) REFERENCES useracct(u\_id)\\
FOREIGN KEY (i\_id) REFERENCES item(i\_id)\\
);\\
A: The SQL query for the question 'How many users who did not leave any review.' needs these table and columns = [USERACCT.name, REVIEW.u\_id], so we need a JOIN operation on the USERACCT and REVIEW tables using the u\_id column because we need to find users who did not leave any review.\\
So we label it as JOIN. Plus, it requires nested queries with (INTERSECT, UNION, EXCEPT, IN, NOT IN), and we need the answer to the questions = ['What is the list of u\_id who left a review?']. So it needs nested queries and we label it as NESTED. \\
Thus the SQL query can be classified as JOIN, NESTED.\\
Label: JOIN, NESTED\\
\\
\#\#\# Issues you should be concerned about:\\
table info:\textbf{[table\_info]}\\
Q: \textbf{[query]}\\
A: \\

\subsection{SQL Generation Prompt}
\label{sec_appendiix:sql_generation}
% \subsubsection{zero-shot}

\subsubsection{Few-shot: easy class}
\textbf{[few-shot]} \\
\#\#\# Database scheme: \textbf{[table\_info]} \\
\#\#\# Please think carefully about the related fields or calculation methods of '\textbf{[main\_metric]}', then write valid SQLite to solve the following questions based on the above table information, and do not select extra columns that are not explicitly requested in the query.\\
\#\#\# Query: \textbf{[query]}\\
\#\#\# specification\\
1.In sql, just select columns that are explicitly requested in the query.\\
2.The output format must strictly meet the given json specification: {{"sql": "ccc"}}

\subsubsection{Few-shot: join class}
\textbf{[few-shot]}\\
\#\#\# Database scheme: \textbf{[table\_info]}\\
\#\#\# Please think carefully about the related fields or calculation methods of '\textbf{[main\_metric]}', then write valid SQLite to solve the following questions based on the above table information, and do not select extra columns that are not explicitly requested in the query.\\
\#\#\# Query: \textbf{[query]}\\
\#\#\# HINT: The question may need connection operation like JOIN.\\
\#\#\# specification\\
1."LIMIT" just is used when explicitly requesting how much to retrieve in the query.\\
2.In sql, just select columns that are explicitly requested in the query.\\
3.The output format must strictly meet the given json specification: {"sql": "ccc"}

\subsubsection{Few-shot: join-nested or nested class}
\textbf{[few-shot]}\\
\#\#\# Database scheme: \\
\textbf{[table\_info]}\\
\#\#\# Please think carefully about the related fields or calculation methods of '\textbf{[main\_metric]}', then write valid SQLite to solve the following questions based on the above table information, and do not select extra columns that are not explicitly requested in the query.\\
\#\#\# Query: \textbf{[query]}\\
\#\#\# HINT: The question may needs nested queries like INTERSECT, UNION, EXCEPT, NOT IN.\\
\#\#\# specification\\
1."LIMIT" just is used when explicitly requesting how much to retrieve in the query.\\
2.Don't use "IN", "OR", "LEFT JOIN" in sql because they aren't supported in execution engine, you can use "INTERSECT" or "EXCEPT" instead.\\
3.In sql, just select columns that are explicitly requested in the query.\\
4.The output format must strictly meet the given json specification: {{"sql": "ccc"}}

\subsubsection{The Specification of Bird Dataset}
1."LIMIT" is only used when explicitly requesting how much to retrieve in the query or for extreme value problems.
2.In sql, just select columns that are explicitly requested in the query.
3.Please note that the required fields in the ORDER BY clause and SELECT require the addition of a WHERE condition IS NOT NULL.
4.Note that a single-word field name does not require double quotes, but a field name composed of multiple words requires double quotes.
5.Pay attention to the division operation. It is necessary to perform a type conversion operation on molecule A, such as: cast(A as REAL).

\subsection{Self-Correction Prompt}
\label{sec_appendiix:self_correction}
For the given question, use the Database scheme to fix the given SQLite QUERY for any issues. \\
If there are any problems, please fix them. \\
If there are no issues, return SQLite QUERY as is. \\
\#\#\# There are some instructions for fixing the SQL query: \\
1) In sql, just select columns that are explicitly requested in the query. \\
2) Pay attention to the columns that are used for the SELECT clause. Fix possible ambiguous columns if there are the same columns in different table in the SELECT clause. \\
3) Pay attention to the correspondence between tables and fields. Cannot use fields that are not in the table. \\
4) Pay attention to the columns that are used for the JOIN. The join table condition must be in the Foreign\_keys. \\
5) Pay attention to the use of the JOIN. Don't use LEFT JOIN unless necessary.  \\
6) Only change the SELECT, GROUP BY and ORDER BY clause when necessary.  \\
7) Database scheme: \textbf{[table\_info]}  \\
\#\#\# Question: \textbf{[query]}  \\
\#\#\# SQLite SQL QUERY:  \\
\textbf{[sql]}  \\
\#\#\# Fixed SQL QUERY:  \\
SELECT  \\

\subsection{Active Learning Prompt}
\label{sec_appendiix:active_learning}

\subsubsection{Activate Learning Prompt of Spider Dataset}
Please determine the type of question. If it is an extremum problem, modify the SQL accordingly.   \\
If not, use the original SQL as the modified SQL.  \\
Q1: What is the name of the instructor who advises the student with the greatest number of total credits?  \\
original SQL: SELECT T2.name FROM instructor T2 JOIN advisor T1 ON T2.id = T1.i\_id JOIN student s ON T1.s\_id = T3.id WHERE T3.tot\_cred = (SELECT MAX(tot\_cred) FROM student)  \\
A: The question is an extremum problem, so i should modify the SQL. 
The modified SQL: SELECT T2.name FROM advisor AS T1 JOIN instructor AS T2 ON T1.i\_id = T2.id JOIN student AS T3 ON T1.s\_id = T3.id ORDER BY T3.tot\_cred DESC LIMIT 1  \\
Q2: Return the id and full name of the customer who has the fewest accounts.  \\
original SQL: SELECT c.customer\_id, c.customer\_first\_name, c.customer\_last\_name FROM CUSTOMERS c JOIN ACCOUNTS a ON c.customer\_id = a.customer\_id GROUP BY c.customer\_id HAVING COUNT(a.account\_id) = (SELECT COUNT(account\_id) FROM ACCOUNTS GROUP BY customer\_id ORDER BY COUNT(account\_id) ASC LIMIT 1)  \\
A: The question is an extremum problem, so i should modify the SQL. 
The modified SQL: SELECT T1.customer\_id, T2.customer\_first\_name, T2.customer\_last\_name FROM Customers\_cards AS T1 JOIN Customers AS T2 ON T1.customer\_id = T2.customer\_id GROUP BY T1.customer\_id ORDER BY count(*) ASC LIMIT 1 \\
Q3: What is the average hours across all projects? \\
original SQL: SELECT avg(hours) FROM projects  \\
A: The question is not an extremum problem, so i should use the original SQL as the modified SQL. \\
The modified SQL: SELECT avg(hours) FROM projects \\
Q4: \textbf{[query]} \\
\textbf{[table\_info]} \\
original SQL: \textbf{[sql]} \\
A: \\

\subsubsection{Activate Learning Prompt of Bird Dataset}
You are an experienced SQL engineer. Please correct the given problem according to the tips and examples given. If it is not within the tips range or there is no issues, return SQLite QUERY as is.\\
Tip1: In sql, just select columns that are explicitly requested in the query.\\
Question: Return the id and first name of the customer who has the fewest accounts.\\
SQLite SQL QUERY: SELECT T1.customer\_id, T2.customer\_first\_name, T2.customer\_last\_name FROM Customers\_cards AS T1 JOIN Customers AS T2 ON T1.customer\_id = T2.customer\_id GROUP BY T1.customer\_id ORDER BY count(*) ASC LIMIT 1\\
Fixed SQL QUERY: SELECT T1.customer\_id, T2.customer\_first\_name FROM Customers\_cards AS T1 JOIN Customers AS T2 ON T1.customer\_id = T2.customer\_id GROUP BY T1.customer\_id ORDER BY count(*) ASC LIMIT 1\\
\\
Tip2: Pay attention to the columns that are used for the SELECT clause. Fix possible ambiguous columns if there are the same columns in different table in the SELECT clause and Do not use `||` to connect multiple fields.\\
Question: Please list the full name of the youngest person\\
SQLite SQL QUERY: SELECT first\_name || ' ' || last\_name AS full\_name FROM member join profile on profile.id = member.id ORDER BY profile.age ASC LIMIT 1\\
Fixed SQL QUERY: SELECT member.first\_name, member.last\_name FROM member join profile on profile.id = member.id ORDER BY profile.age ASC LIMIT 1\\
\\
Tip3: Pay attention to standard writing for optimal value problems.\\
Question:  State the most popular movie? \\
HINT: most popular movie refers to MAX(movie\_popularity);\\
SQLite SQL QUERY: SELECT movie\_title FROM movies WHERE movie\_popularity = (SELECT MAX(movie\_popularity) FROM movies);\\
Fixed SQL QUERY: SELECT movie\_title FROM movies ORDER BY movie\_popularity DESC LIMIT 1\\
\\
Tip4: Note that the joint table condition must be in the Foreign\_key.\\
Foreign\_key: [lists.user\_id = lists\_users.user\_id, ratings.movie\_id = movies.id, link.id = movies.id];
Question: What is the average rating for movie titled 'When Will I Be Loved'?\\
SQLite SQL QUERY: SELECT avg(ratings.rating\_score) FROM movies INNER JOIN ratings ON movies.movie\_id = ratings.movie\_id WHERE movies.movie\_title = 'When Will I Be Loved'\\
Fixed SQL QUERY: SELECT avg(ratings.rating\_score) FROM movies INNER JOIN ratings ON movies.id = ratings.movie\_id WHERE movies.movie\_title = 'When Will I Be Loved'\\
The following is the question that requires you to decide whether to make changes\\
Database scheme: \textbf{[table\_info]}\\
Question: \textbf{[query]}\textbf{[hint]}\\
SQLite SQL QUERY: \textbf{[sql]} \\
Fixed SQL QUERY: SELECT\\

\end{document}